\pdfoutput=1

\documentclass[11pt]{article}

\usepackage{ACL2023}

\usepackage{times}
\usepackage{latexsym}
\usepackage{graphicx}
\usepackage[T1]{fontenc}

\usepackage[utf8]{inputenc}
\usepackage{makecell}
\usepackage{microtype}

\usepackage{stfloats}
\usepackage{inconsolata}
\usepackage{amssymb}
\usepackage{subcaption}
\usepackage{amsmath}
\usepackage{cleveref}
\usepackage{bm}
\usepackage{float}
\usepackage{pifont}
\usepackage{booktabs}
\usepackage{multirow}
\usepackage{enumitem}

\usepackage{algorithm}
\usepackage{CJKutf8}

\usepackage{algpseudocode}

\title{LaCo: Large Language Model Pruning via Layer Collapse}

\author{Yifei Yang$^{1,2,3}$, Zouying Cao$^{1,2,3}$, Hai Zhao$^{1,2,3}$\thanks{\,\, Corresponding author. This research was supported by the Joint Research Project of Yangtze River Delta Science and Technology Innovation Community (No. 2022CSJGG1400),
the Joint Funds of the National Natural Science Foundation of China (Grant No. U21B2020).} \\
$^1$Department of Computer Science and Engineering, Shanghai Jiao Tong University\\
$^2$Key Laboratory of Shanghai Education Commission for Intelligent Interaction \\
and Cognitive Engineering, Shanghai Jiao Tong University \\
$^3$Shanghai Key Laboratory of Trusted Data Circulation and Governance in Web3 \\
\texttt{\{yifeiyang,zouyingcao\}@sjtu.edu.cn, zhaohai@cs.sjtu.edu.cn} \\}

\begin{document}
\maketitle
\begin{abstract}
Large language models (LLMs) based on transformer are witnessing a notable trend of size expansion, which brings considerable costs to both model training and inference. However, existing methods such as model quantization, knowledge distillation, and model pruning are constrained by various issues, including hardware support limitations, the need for extensive training, and alterations to the model internal structure. In this paper, we propose a concise layer-wise structured pruner called \textit{Layer Collapse (LaCo)}, in which rear model layers collapse into a prior layer, enabling a rapid reduction in model size while preserving the model structure. Comprehensive experiments show that our method maintains an average task performance of over 80\% at pruning ratios of 25-30\%, significantly outperforming existing state-of-the-art structured pruning methods. We also conduct post-training experiments to confirm that the \textit{LaCo} effectively inherits the parameters of the original model. Additionally, we perform ablation studies on various settings of \textit{LaCo}. Finally, we discuss our motivation from the perspective of layer-wise similarity and evaluate the performance of the pruned LLMs across various pruning ratios\footnote{\url{https://github.com/yangyifei729/LaCo}}.
\end{abstract}

\section{Introduction}

Recently, large language models (LLMs) based on Transformer~\citep{vaswani2017attention} have showcased impressive capabilities across diverse tasks. However, the prevailing trend in model development leans towards larger scales, placing substantial demands on computational resources.

To mitigate the above challenge, various approaches have been explored to reduce the inference and training costs of models or to derive a smaller model from an LLM, including model quantization \cite{dettmers2022llm,yao2022zeroquant,xiao2023smoothquant}, knowledge distillation \cite{liu2022multi,hsieh2023distilling,shridhar2023distilling}, and model pruning \cite{zhang2022platon,frantar2023sparsegpt,ma2023llm}. However, existing solutions exhibit certain notable drawbacks. Model quantization typically necessitates specific hardware support and often impacts model performance. Knowledge distillation often requires retraining a smaller model, which is costly and task-specific. Model pruning, whether non-structured or structured, has its issues. Non-structured pruning often involves model sparsity, which generally leads to certain performance loss and also relies on hardware support. Structured pruning entails removing specific modules, often altering the model structure and diminishing the model portability.

Considering the above issues, we contemplate directly pruning the model with a new idea: to prune some layers directly from a well-trained LLM and substitute the parameters of one layer for multiple layers, enabling effective model pruning.

\begin{figure}[!tp]
    \centering    \includegraphics[width=0.98\linewidth,scale=1.00]{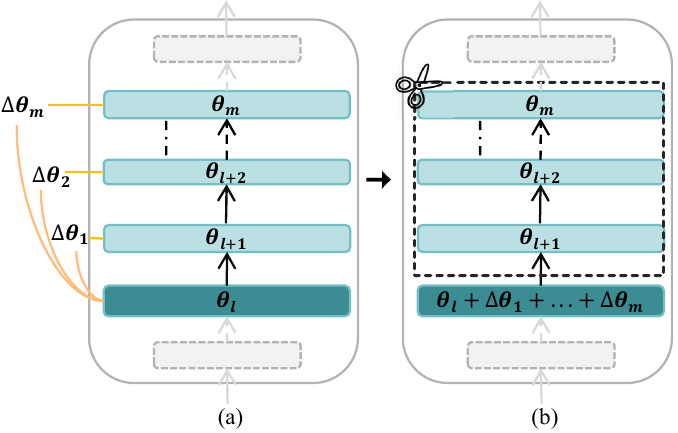}
    \caption{An example of \textit{Reserving-Differences-while-Seeking-Common (RDSC) Layer Merge}. In (a), we perform parameter differencing, which we regard as Reserving-Differences. In (b), we conduct parameter merging, which we interpret as Seeking-Common.}
    \label{fig:diff_merge}
\end{figure}





Specifically, we observe that merging the parameter differentials of certain layers with their subsequent layers often does not significantly impact model performance, as illustrated in Figure~\ref{fig:diff_merge}. 

We term it the \textit{Reserving-Differences-while-Seeking-Common (RDSC) Layer Merge}, as it incorporates parameter differencing and merging. Building upon this insight, we introduce a streamlined yet potent layer-wise pruner dubbed \textit{Layer Collapse (LaCo)}, in which rear layers collapse into a prior layer, with the objective of preserving the model's output representation. In this paper:

$\bullet$ The \textit{Layer Collapse} can directly remove 30\%-50\% of model layers without training while maintaining the model performance. Experiments on multiple benchmarks show that our approach outperforms state-of-the-art structured pruning methods under equivalent pruning ratios.

$\bullet$ The \textit{Layer Collapse} preserves the internal structure of LLMs, such as maintaining intermediate dimensions. So, the pruned models can be seamlessly integrated into existing applications without any changes to the system's implementation.

$\bullet$ We conduct post-training to confirm that \textit{Layer Collapse} can efficiently inherit parameters and requires only minimal training to restore the pruned model to the original model's loss convergence level. Additionally, we discuss our motivation and evaluate the performance of pruned models using \textit{LaCo} across different pruning ratios. We also perform ablation studies on various settings of \textit{LaCo}.

\section{Method}
\subsection{Reserving-Differences-while-Seeking-Common Layer Merge}
For the $l$-th layer of an LLM, we denote all its parameters, including those in self-attention (SAN) and MLP as $\boldsymbol{\theta}_{l}$. For the $m$ consecutive layers following it, we merge the parameters of $\boldsymbol{\theta_{l+1}}, \boldsymbol{\theta_{l+2}}, \cdots, \boldsymbol{\theta_{l+m}}$ into $\boldsymbol{\theta_{l}}$ to form $\boldsymbol{\theta^{*}_{l}}$:

\begin{equation}\label{eq:merge}
  \begin{aligned}
    \boldsymbol{\theta}^{*}_{l} &= \boldsymbol{\theta}_{l} + (\boldsymbol{\theta}_{l+1}-\boldsymbol{\theta}_{l}) + \cdots + (\boldsymbol{\theta}_{l+m}-\boldsymbol{\theta}_{l}) \\
    &= \boldsymbol{\theta}_{l} + \sum_{k=1}^{m}(\boldsymbol{\theta}_{l+k} - \boldsymbol{\theta}_{l})
  \end{aligned}
\end{equation}
where $(\boldsymbol{\theta_{l+k}} - \boldsymbol{\theta_{l})}$ is the layer-wise parameter difference. Given identical layer structures, we independently apply these processes to both SAN and MLP. Then, these $m$ consecutive layers will be discarded. Subsequent model pruning will continuously involve RDSC Layer Merge which can be regarded as the continual collapse of layers onto specific layers, hence the name \textit{Layer Collapse}.

\begin{figure*}[!tp]
    \centering
    \includegraphics[width=0.95\linewidth,scale=1.00]{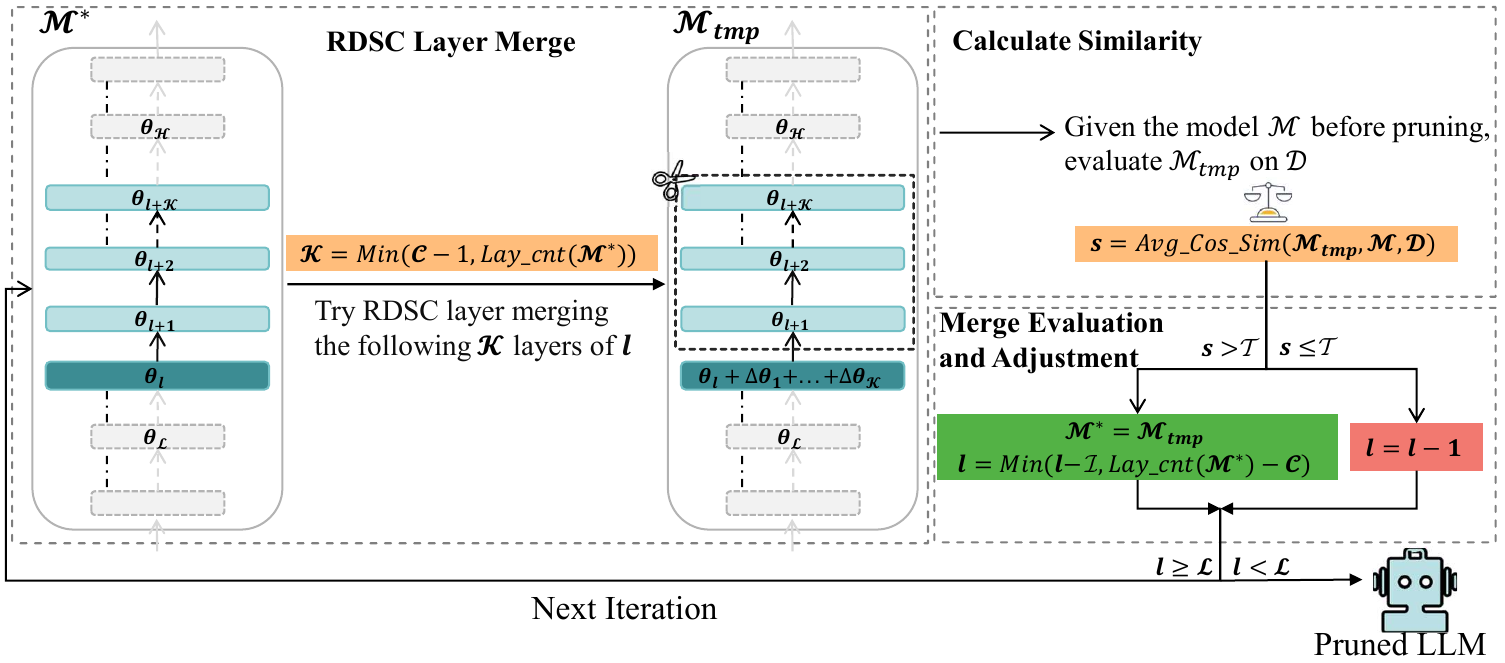}
    \caption{An illustration of Layer Collapse.}
    \label{fig:main_method}
\end{figure*}

\begin{algorithm}[!tb]
\caption{Workflow of Layer Collapse}
\label{algor:pruning}
\begin{algorithmic}[1]
    \Require LLM $\mathcal{M}$; Number of layers combined in each merge $\mathcal{C}$; Layer range $[\mathcal{L}, \mathcal{H}]$; Minimum interval between two adjacent merged layers $\mathcal{I}$; Few-shot Calibration Samples $\mathcal{D}$; Threshold for representation similarity $\mathcal{T}$
        
    \Ensure Pruned LLM $\mathcal{M}^{*}$

    \State $\mathcal{M}^{*} \gets \mathcal{M}$
    \State $l \gets \mathcal{H} - \mathcal{C}$
    \While{$l >=\mathcal{L}$}

        \State $\mathcal{K} \gets$ Min($\mathcal{C}-1$, Layer\_Count($\mathcal{M}^{*}$)$-l$)
        
        \State $\mathcal{M}_{tmp} \gets $ RDSC\_Lay\_Merge($\mathcal{M}^{*}$, $l$, $\mathcal{K}$)
        
        \State $s\gets$ Avg\_Cos\_Sim($\mathcal{M}_{tmp}$, $\mathcal{M}$, $\mathcal{D}$)
        \If{$s > \mathcal{T}$}
            \State $\mathcal{M}^{*} \gets \mathcal{M}_{tmp}$
            \State $l \gets l-\mathcal{I}$
            \If{$l>$ Layer\_Count($\mathcal{M}^{*}$)}
                \State $l \gets $Layer\_Count($\mathcal{M}^{*}$)$ - \mathcal{C}$

            \EndIf
        \Else
            \State $l\gets l-1$
        \EndIf
    \EndWhile
    \State \Return $\mathcal{M}^{*}$

\end{algorithmic}
\end{algorithm}

\subsection{Layer Collapse}
We dynamically merge adjacent layers from the topmost layer down, ensuring the pruned model's output representation on few-shot calibration samples remains as similar as possible to the original model to minimize performance loss. Algorithm \ref{algor:pruning} summarizes the workflow of \textit{Layer Collapse}:

\noindent \textbf{(1) Preparation}

For an LLM $\mathcal{M}$ to be pruned, we define the number of layers to be merged during each merging operation as $\mathcal{C}$. We configure the merging to operate within a certain range of layers, denoted as $[\mathcal{L}, \mathcal{H}]$. As the layer merging operation inevitably leads to a performance loss, to prevent consecutive layer merging from causing a sharp decline in the model performance, we set a minimum interval of layers between two merging operations as $\mathcal{I}$. Few-shot calibration samples $\mathcal{D}$, typically a few plain sentences, are used during the pruning process. We perform forward computations on $\mathcal{D}$ with both the pruned and original models to obtain the output representations and ensure that the similarity of representations is not less than the threshold $\mathcal{T}$.

\noindent \textbf{(2) Pruning (line 1-17)}

We present an illustration of \textit{Layer Collapse} in Figure~\ref{fig:main_method}. We begin by initializing the model $\mathcal{M}^{*}$ with the model $\mathcal{M}$ and set a layer pointer $l$ to start from $\mathcal{H} - \mathcal{C}$. Then, the iterative process begins:

\textbf{\textit{RDSC Layer Merge (line 4-5)}} During each iteration, our approach involves merging the $\mathcal{K}$ layers following layer $l$ into layer $l$ itself and then discarding the redundant $\mathcal{K}$ layers, where $\mathcal{K}$ is the minimum of $\mathcal{C}-1$ and the total layer count of $\mathcal{M}^{*} - l$, implying merging either the subsequent $\mathcal{C}-1$ layers or all layers following $l$, thus to prune the model $\mathcal{M}^{*}$, resulting in the interim model $\mathcal{M}_{tmp}$.

\textbf{\textit{Calculate Similarity (line 6)}} We process each sentence in $\mathcal{D}$ using forward computations with $\mathcal{M}_{tmp}$ and $\mathcal{M}$ to derive their representations which are the output hidden-states of the last layer of the model. For every sentence, we then calculate the cosine similarity between these representations from both models, averaging these values to obtain the overall similarity score $s$.

\textbf{\textit{Merge Evaluation and Adjustment (line 7-15)}} Then, we evaluate $s$ against the threshold $\mathcal{T}$. Should $s$ exceed $\mathcal{T}$, the current merge is considered successful. Then,  $\mathcal{M}_{tmp}$ is updated to $\mathcal{M}^{*}$ for the next iteration, and the pointer $l$ is adjusted downwards by $\mathcal{I}$ layers. Conversely, $l$ is simply reduced by a single layer. It is important to highlight that the instances may occur where $l$ falls below the total layer count of $\mathcal{M}^{*}$ after a series of successive merges. Consequently, it is required to reset $l$ to the layer count in $\mathcal{M}^{*} - \mathcal{C}$, as illustrated in line 11.

We iterate through the above process until $l$ is less than $\mathcal{L}$ and output the pruned LLM.

\subsection{Complexity Analysis}
The complexity of \textit{LaCo} primarily depends on model inference. In the worst-case scenario, with $\mathcal{L}$ set to 0 and $\mathcal{H}$ to the total number of layers, if in each iteration the similarity $s$ is less than $\mathcal{T}$, all layers will be traversed. Thus, the worst-case time complexity is $O(\mathcal{H}\times||\mathcal{D}||)$. For example, for Llama2-13B with 40 layers and $||\mathcal{D}||$ consisting of 10 sentences, the maximum number of inference steps would be only 400 sentences, which can be completed within minutes on a single GPU.

\section{Experiments}
\subsection{Models}
To assess the effectiveness of the proposed \textit{LaCo}, we conduct experiments on popular English LLMs, Llama2-7B and 13B \cite{touvron2023llama}. Additionally, we test the effectiveness on bilingual LLMs, specifically Baichuan2-7B and 13B \cite{yang2023baichuan}, which support both Chinese and English. We leverage the base versions of these LLMs. All these models are decoder-only models based on the transformer architecture.

\subsection{Benchmarks}

To comprehensively evaluate the pruned model's capabilities, we utilized the OpenCompass evaluation framework \cite{2023opencompass}. Specifically, following OpenCompass categorization, we conduct evaluations in five aspects: Reasoning, Language, Knowledge, Examination and Understanding. We select several benchmarks from each category.
\textbf{Reasoning:} CMNLI \cite{xu2020clue}, HellaSwag (HeSw) \cite{zellers2019hellaswag}, PIQA \cite{bisk2019PIQA}.
\textbf{Language:} CHID \cite{zheng-etal-2019-chid}, WSC \cite{levesque2012winograd}.
\textbf{Knowledge:} CommonSenseQA (CSQA) \cite{talmor2018commonsenseqa}, BoolQ \cite{clark2019boolq}.
\textbf{Examination:} MMLU \cite{hendryckstest2021}, CMMLU \cite{li2023cmmlu}.
\textbf{Understanding:} Race-High/Middle (H/M) \cite{lai2017race}, XSum \cite{narayan2018dont}, C3 \cite{sun2019investigating}.

We conduct evaluations using official scripts from OpenCompass, all zero-shot or few-shot, without additional training. Two evaluation modes are utilized: perplexity (PPL) and generation (GEN)~\footnote{opencompass.readthedocs.io/en/latest/get\_started/faq.html}. For CHID and XSum, we use the GEN mode. For the WSC dataset, we use both PPL (WSC$_{\text{P}}$) and GEN (WSC$_{\text{G}}$) modes. The remaining benchmarks are evaluated using the PPL mode. The evaluation results on each benchmark are converted to a score by OpenCompass, where a higher score indicates better performance. OpenCompass provides official evaluation results for the Baichuan2 and Llama2 series. However, to avoid discrepancies resulting from hardware and software environments, as well as potential errors in official results, we reproduce all results to ensure fairness.

\subsection{Baselines}
Since \textit{LaCo} involves structured pruning, which directly removes components from LLMs, we select two state-of-the-art (SOTA) structured pruning methods, LLM-Pruner (LLMPru.)~\cite{ma2023llm} and SliceGPT~\cite{ashkboos2024slicegpt}, as our baselines. These methods have surpassed the previous SOTA sparsity method, SparseGPT~\cite{frantar2023sparsegpt}. In our experiments, we set the pruning ratios of baselines to be equivalent to or slightly smaller than \textit{LaCo} to ensure fairness.

\subsection{Settings}

Since previous work mostly set pruning ratios below 30\%, we heuristically adjust the hyperparameters to bring the model pruning ratio close to 30\%, as shown in Appendix~\ref{app:hyper_setting} Table~\ref{table:hparams}. We randomly select 5 sentences from both the English and Chinese Wikipedia datasets for Baichuan2 and 10 sentences from English Wikipedia for Llama2 as few-shot calibration samples. All experiments are conducted on a server with 8 Nvidia A100 80GB GPUs.

\begin{table*}[!htp]
\setlength\tabcolsep{0.8pt} 
\small
    \centering
    \resizebox{0.97\linewidth}{!}{
    \begin{tabular}{c|c|c|ccc|ccc|cc|cc|cccc}
    \toprule[0.5pt]
    \midrule
    \multirow{2}{*}{LLM} & \multirow{2}{*}{Pruner} & \multirow{2}{*}{Ratio/Lay.} & \multicolumn{3}{c|}{Reasoning} & \multicolumn{3}{c|}{Language} & \multicolumn{2}{c|}{Knowledge} & \multicolumn{2}{c|}{Examination} & \multicolumn{4}{c}{Understanding} \\

    & &  & CMNLI & HeSw & PIQA & CHID & WSC$_{\text{P}}$ & WSC$_{\text{G}}$ & CSQA & BoolQ & MMLU & CMMLU &  Race$_{\text{H}}$ &  Race$_{\text{M}}$ & XSum & C3 \\

    \midrule
 
    \multirow{4}{*}{\makecell[c]{\textbf{Llama2} \\ \textbf{-7B}}} & Dense & 0\%/32 & 34.90 & 74.00  & 78.30  & 46.50   &  - & 66.30 & 66.50  & 74.90   & 46.80   & 31.80 & 37.50 & 40.20 & 19.70 & 42.80 \\
            \cmidrule{2-17}
            & Dense$^{*}$ & 0\%/32 & 32.98 & 71.35 &  78.18 & 46.04 & 37.50 & 38.46 & 66.67 & 70.67 & 45.92 & 31.86 & 35.51 & 33.15 & 19.68 & 43.78   \\
            \cmidrule{2-17}
            & LLMPru.  & 27.0\%/32 & 34.33 & \textbf{56.46} & \textbf{71.22} & 25.25 & 36.54 & 0.96 & 42.51 & 55.20 & 23.33 & 25.25 & 22.56 & 22.35 & 11.51 & 25.64\\
            \cmidrule{2-17}
            & SliceGPT  & 26.4\%/32 & 31.70 & 50.27 & 66.21 & 20.79 & 36.54 & 19.23 & 41.36 & 38.32 & \textbf{28.92} & \textbf{25.37} & 21.07 & 21.66 & 4.89 & \textbf{39.78} \\
            \cmidrule{2-17}
            & \textbf{LaCo}  & \textbf{27.1\%/23} & \textbf{34.43} & 55.69 & 69.80 & \textbf{36.14} & \textbf{40.38} & \textbf{25.00} & \textbf{45.70} & \textbf{64.07} & 26.45 & 25.24 & \textbf{22.61} & \textbf{23.61} & \textbf{15.64} & 39.67 \\
    \midrule
    \multirow{4}{*}{\makecell[c]{\textbf{Llama2} \\ \textbf{-13B}}} & Dense & 0\%/40 & 41.40 & 77.50 & 79.80 & 53.00 & - & 66.30 & 66.70 & 82.40 & 55.00 & 38.40 & 58.90 & 63.00 & 23.40 & 46.10 \\
            \cmidrule{2-17}
            & Dense$^{*}$  & 0\%/40 & 32.99 & 74.83 & 79.71 & 52.97 & 50.96 & 63.46 & 66.91 & 71.50 & 55.63 & 38.74 & 58.03 & 60.24 & 23.56 & 47.51 \\
            \cmidrule{2-17}
            & LLMPru. & 24.4\%/40 & \textbf{33.03} & \textbf{67.76} & \textbf{76.66} & 35.64 & 40.38 & 0.00 & 50.86 & 56.42 & 25.21 & 24.71 & 22.47 & 22.08 & \textbf{19.17} & 32.33\\
            \cmidrule{2-17}
            & SliceGPT  & 23.6\%/40 & 29.82 & 55.71 & 69.04 & 19.31 & 36.54 & \textbf{36.54} & 47.26 & 37.86 & 37.14 & 25.79 & 23.41 & 24.03 & 5.27 & 41.92 \\
            \cmidrule{2-17}
            & \textbf{LaCo}  & \textbf{24.6\%/30} & 32.86 & 64.39 & 74.27 & \textbf{40.10} & \textbf{52.88} & 35.58 & \textbf{52.66} & \textbf{63.98} & \textbf{45.93} & \textbf{32.62} & \textbf{54.49} & \textbf{56.55} & 14.45 & \textbf{44.93} \\
    \midrule
    \multirow{4}{*}{\makecell[c]{\textbf{Baic2.} \\ \textbf{-7B}}} & Dense & 0\%/32 & 32.90 & 67.00 & 76.20 & 82.70 & - & 66.30 & 63.00 & 63.20 & 54.70 & 57.00 & 52.50 & 50.90 & 20.90 & 64.50 \\
            \cmidrule{2-17}
            & Dense$^{*}$  & 0\%/32 & 33.37 & 67.56 & 76.17 & 82.67 & 41.35 & 63.46 & 63.14 & 63.30 & 54.25 & 56.95 & 52.63 & 51.04 & 20.84 & 64.55 \\
            \cmidrule{2-17}
            & LLMPru. & 24.2\%/32 & 32.28 & \textbf{53.66} & \textbf{71.82} & 69.80 & \textbf{53.85} & 0.00 & \textbf{47.83} & \textbf{61.19} & 24.93 & 25.69 & 21.96 & 22.28 & \textbf{15.98} & 41.64 \\
            \cmidrule{2-17}
            & SliceGPT  & 22.2\%/32 & 32.07 & 25.29 & 50.33 & 14.85 & 36.54 & 0.00 & 19.57 & 39.30 & 25.18 & 25.25 & 23.53 & 22.49 & 0.00 & 26.58 \\
            \cmidrule{2-17}
            & \textbf{LaCo}  & \textbf{24.2\%/23} & \textbf{33.00} & 52.28 & 68.50 & \textbf{76.24} & 42.31 & \textbf{26.92} & 47.26 & 56.15 & \textbf{31.53} & \textbf{31.24} & \textbf{28.99} & \textbf{27.72} & 12.03 & \textbf{50.85} \\
    \midrule
    \multirow{4}{*}{\makecell[c]{\textbf{Baic2.} \\ \textbf{-13B}}} & Dense & 0\%/40 & 32.70 & 70.80 & 78.10 & 83.20 & - & 63.20 & 65.60 & 67.00 & 59.50 & 61.30 & 67.20 & 68.90 & 25.20 & 65.60 \\
            \cmidrule{2-17}
            & Dense$^{*}$  & 0\%/40 & 33.21 & 71.10 & 78.07 & 83.17 & 41.35 & 63.46 & 65.60 & 67.00 & 58.81 & 61.27 & 67.27 & 68.94 & 24.95 & 65.64 \\
            \cmidrule{2-17}
            & LLMPru. & 24.3\%/40 & \textbf{33.80} & 53.57 & \textbf{71.82} & 72.77 & 37.50 & 0.00 & 38.82 & 56.54 & 23.19 & 25.18 & 21.17 & 21.61 & \textbf{13.67} & 39.89 \\
            \cmidrule{2-17}
            & SliceGPT  & 22.8\%/40 & 32.07 & 25.85 & 51.03 & 10.40 & 36.54 & 0.00 & 18.02 & 37.83 & 22.95 & 25.26 & 21.56 & 21.52 & 0.00 & 24.99 \\
            \cmidrule{2-17}
            & \textbf{LaCo}  & \textbf{24.7\%/30} & 33.03 & \textbf{60.71} & 68.88 & \textbf{76.73} & \textbf{44.23} & \textbf{60.58} & \textbf{55.45} & \textbf{62.35} & \textbf{51.35} & \textbf{53.65} & \textbf{56.92} & \textbf{57.80} & 12.32 & \textbf{61.10} \\
    \midrule
    \bottomrule
    \end{tabular}"}
    \caption{The main results of our experiments. \textit{Lay.} is the number of model layers. \textit{Dense} is the official LLM results in OpenCompass and \textit{Dense$^{*}$} is our reproduction. \textit{LLMPru.} and \textit{SliceGPT} are two baseline comparisons.}
    \label{tab:main_res}
\end{table*}

\subsection{Main Results}
In Table~\ref{tab:main_res}, we present the results of four LLMs under different pruning methods across various benchmarks. ``Dense'' represents the official results of the unpruned LLMs in OpenCompass leaderboards, while ``Dense$^{*}$'' represents our reproduction of the ``Dense'' results. "LLMPru." and "SliceGPT" correspond to the two baselines, respectively. ``Ratio" refers to the overall pruning ratio, namely the proportion of the total number of pruned parameters to that of the unpruned model. ``Lay.'' denotes the total number of layers in the model.

Comparing Dense and Dense$^{*}$, the results show not much difference, with most discrepancies within 5\%. This indicates our experimental setup is error-free. To ensure fairness, we compare the results against Dense$^{*}$ in the subsequent analyses.

Upon comparing LaCo with the baselines, from Table~\ref{tab:main_res}, it can be observed that LaCo achieves the best results on most benchmarks, despite our pruning ratio being slightly higher than the baselines. 

To provide a more intuitive presentation of the results in Table~\ref{tab:main_res}, we compute the average scores of each pruner across all benchmarks (Avg.), the average scores per category (Reas., Lan., Know., Exam., Unde.), and the average performance percentages relative to Dense$^{*}$ across all benchmarks (Per.) in Table~\ref{tab:avg_res}. Overall, our average scores are significantly higher than the baselines. LaCo shows superior performance in four out of five categories. Though there is a slight dip in Reasoning, it remains comparable. Additionally, LaCo's average performance percentage across all datasets, relative to Dense$^{*}$, is far superior to the baselines. The average percentage surpasses 80\% in three out of four models, with the lowest being 73\% on Baichuan2-7B. In contrast, none of the baselines exceed 70\%.

To demonstrate the stability of the pruned models by LaCo, we compute the performance percentage relative to Dense$^{*}$  (Appendix~\ref{app:perf_prec} Table~\ref{tab:main_res_per}). LaCo-pruned models maintain performance above 70\% on most benchmarks and do not experience crashes, with no performance dropping below 30\%.

Notably, on three benchmarks evaluated through GEN mode, CHID, XSUM, and WSC$_G$, the LLMs pruned by LaCo maintain relatively stable performance, while models pruned by baselines exhibit poorly, with even multiple results becoming 0.00. GEN mode scores are based on the model's generated sentences, and the models pruned by baselines are prone to producing meaningless repetitive outputs. In Appendix~\ref{app:ex_res} Table~\ref{tab:response_dif_pruner}, we showcase an example from the Xsum benchmark, where Llama2-7B, pruned by baselines, produces nonsensical repeated outputs, whereas our LaCo yields outputs resembling normal sentences.

We also conduct experiments with Llama2-70B on several benchmarks. The results in Appendix~\ref{app:exp_llama_70} Table~\ref{tab:exp_70} show that LaCo still outperforms the baseline on larger-scale model.

In summary, LaCo is a superior pruner that preserves model performance and demonstrates exceptional stability across various benchmarks. It relies solely on parameter differences and additions, without altering the model's internal structure, resulting in a concise and efficient pruning solution.

\begin{table}[!tp]
\setlength\tabcolsep{0.6pt} 
\small
    \centering
    \resizebox{0.95\linewidth}{!}{
    \begin{tabular}{c|c|c|c|c|c|c|c|c}
    \toprule[0.5pt]
    \midrule
    LLM & Pruner & Avg. & Per. & Reas. & Lan. & Know. & Exam. & Unde. \\
    \midrule
    \multirow{4}{*}{\makecell[c]{\textbf{Llama2} \\ \textbf{-7B}}} & Dense$^{*}$ & 46.55 & 100\% & 60.83 & 40.67 & 68.67 & 38.89 & 33.03 \\
            \cmidrule{2-9}
            & LLMPru.  & 32.36 & 67.79\% & \textbf{54.00} & 20.92 & 48.86 & 24.29 & 20.52 \\
            \cmidrule{2-9}
            & SliceGPT  & 31.87 & 67.37\% & 49.39 & 25.52 & 39.84 & \textbf{27.15} & 21.85 \\
            \cmidrule{2-9}
            & \textbf{LaCo}  & \textbf{37.46} & \textbf{80.28\%} & 53.30 & \textbf{33.84} & \textbf{54.89} & 25.85 & \textbf{25.38} \\
    \midrule
    \multirow{4}{*}{\makecell[c]{\textbf{Llama2} \\ \textbf{-13B}}} & Dense$^{*}$ & 55.50 & 100\% & 62.51 & 55.80 & 69.20 & 47.18 & 47.34  \\
            \cmidrule{2-9}
            & LLMPru. & 36.19 & 65.87\% & \textbf{59.15} & 25.34 & 53.64 & 24.96 & 24.01 \\
            \cmidrule{2-9}
            & SliceGPT & 34.97 & 61.78\% & 51.52 & 30.80 & 42.56 & 31.46 & 23.66  \\
            \cmidrule{2-9}
            & \textbf{LaCo}  &  \textbf{47.55} & \textbf{85.21\%} & 57.17 & \textbf{42.85} & \textbf{58.32} & \textbf{39.28} & \textbf{42.60} \\
    \midrule
    \multirow{4}{*}{\makecell[c]{\textbf{Baic2.} \\ \textbf{-7B}}} & Dense$^{*}$ & 56.52 & 100\% & 59.03 & 62.49 & 63.22 & 55.60 & 47.26  \\
            \cmidrule{2-9}
            & LLMPru.  & 38.78 & 69.65\% & \textbf{52.59} & 41.22 & \textbf{54.51} & 25.31 & 25.46 \\
            \cmidrule{2-9}
            & SliceGPT  & 24.36 & 44.27\% & 35.90 & 17.13 & 29.44 & 25.22 & 18.15 \\
            \cmidrule{2-9}
            & \textbf{LaCo}  & \textbf{41.79} & \textbf{73.26\%} & 51.26 & \textbf{48.49} & 51.70 & \textbf{31.38} & \textbf{29.90} \\
    \midrule
    \multirow{4}{*}{\makecell[c]{\textbf{Baic2.} \\ \textbf{-13B}}} & Dense$^{*}$ & 60.70 & 100\% & 60.79 & 62.66 & 66.30 & 60.04 & 56.70  \\
            \cmidrule{2-9}
            & LLMPru.  & 36.40 & 60.70\% & 53.06 & 36.76 & 47.68 & 24.18 & 24.08 \\
            \cmidrule{2-9}
            & SliceGPT  & 23.43 & 40.33\% & 36.32 & 15.65 & 27.92 & 24.10 & 17.02  \\
            \cmidrule{2-9}
            & \textbf{LaCo}  & \textbf{53.94} & \textbf{87.94\%} & \textbf{54.21} & \textbf{60.51} & \textbf{58.90} & \textbf{52.50} & \textbf{47.04} \\
    \midrule
    \bottomrule
    \end{tabular}}
    \caption{The average scores and the percentages comparison with the Dense$^{*}$.}
    \label{tab:avg_res}
\end{table}

\subsection{Comparison of Perplexity}

\begin{table}[!h]
\small
    \centering
    \begin{tabular}{c|c|c|c|c}
    \toprule[0.5pt]
    \midrule
     Pruner & Dense & LaCo & LLM-Pruner & SliceGPT \\
    \midrule
    PPL & 4.46 & \textbf{13.93} & 17.30
 & 14.51 \\
    \midrule
    \bottomrule
    \end{tabular}
    \caption{PPL for different pruners.}
    \label{tab:dif_ppl}
\end{table}

Since PPL is also a commonly used metric for evaluating model performance, we want to compare how LaCo differs from the baseline in maintaining the model's PPL. We evaluate the PPL of the Llama2-7B model with 27\% sparsity using different pruners. The evaluation is performed on 500 sentences selected from Wikipedia, each with a fixed length of 512 tokens.

The results in Table~\ref{tab:dif_ppl} show that the model pruned by LaCo also achieves a lower PPL compared to other baselines, further highlighting the advantage of our LaCo.

\subsection{Pruning Time}

\begin{table}[!htbp]
\small
    \centering
    \begin{tabular}{c|c|c|c}
    \toprule[0.5pt]
    \midrule
     Pruner & LaCo & LLM-Pruner & SliceGPT \\
    \midrule
    Pruning Time & \textbf{14.7s} & 15.9s & 313s \\
    \midrule
    \bottomrule
    \end{tabular}
    \caption{Pruning time for different pruners.}
    \label{tab:pruning_speed}
\end{table}

To verify that LaCo has lower time complexity and faster pruning speed than the baselines, we compare LaCo with them for 27\% sparsity pruning of Llama2-7B on a single A100 GPU. For fairness, we only measure the main pruning process, excluding the time for loading models, loading data, and storing models. The results in Table~\ref{tab:pruning_speed} show LaCo pruning is more efficient compared to the baselines.

\subsection{Memory Usage and Inference Speed}
We also aim to investigate whether the model pruned by LaCo offers advantages in memory usage and inference speed compared to the models pruned by the baselines. In Table~\ref{tab:mem_infer}, we present the average memory consumption and inference speed of the Llama2-13B pruned models from Table~\ref{tab:main_res} on the English Wiki dataset (The results for all models are in Appendix~\ref{app:full_mem_infer} Table~\ref{tab:all_mem_infer}). All models are loaded in Bf16 on a single A100 GPU.

\begin{table}[!htbp]
\small
    \centering
    \begin{tabular}{c|c|c|c|c}
    \toprule[0.5pt]
    \midrule
     Pruner & LaCo & Dense & LLMPru. & SliceGPT \\
    \midrule
    Memory & \textbf{19422} & 25902 & 19874 & 22506 \\
    \midrule
    Infer. & \textbf{38.65} & 29.98 & 27.15 ($\downarrow$) & 35.16 \\
    \midrule
    \bottomrule
    \end{tabular}
    \caption{Memory usage (MB) and inference speed (tokens/s) of the Llama2-13B pruned by different pruners. $\downarrow$ indicates performance worse than the Dense model.}
    \label{tab:mem_infer}
\end{table}

The results indicate that the LaCo-pruned models consume less memory and achieve faster inference speeds. Moreover, while existing baselines may decrease inference speeds compared to the dense model, LaCo does not have this issue.

\section{Further Analysis}
\subsection{Post-training and Re-pruning}
\subsubsection{Post-training}

Due to the inevitable performance loss caused by pruning, we investigate whether models pruned using our LaCo can effectively inherit parameters from the original model and quickly recover performance through post-training on the full parameters. Specifically, we select the pruned Llama2-7B and Baichuan2-7B models obtained through LaCo in the main experiments and post-train them. For training pruned Llama2-7B, we utilize approximately 1.0 billion tokens from the English dataset, while for pruned Baichuan2-7B, we employ approximately 1.25 billion tokens, with a 50\% from English and Chinese. The detailed implementation can be found in the Appendix~\ref{app:impl_det}.

In Figure~\ref{fig:post_train}, we present the loss curves. It can be observed that both models converge rapidly during training, with the loss showing a sharp decline after about 250 steps, then stabilizing. The pruned Llama2-7B and Baichuan2-7B models, both approximately 5 billion parameters, exhibit final convergence losses around 1.6 and 2.0, which are quite comparable to the reported values of 1.75 for Llama2-7B and 1.9 for Baichuan2-7B in their technical reports. The post-training of pruned Llama2-7B and Baichuan2-7B on 4 Nvidia A100 80GB GPUs takes approximately 28 hours and 35 hours, respectively. Training a 5B LLM from scratch requires at least 500 billion tokens on hundreds of A100 GPUs for several months. However, we achieve a loss-converged model of similar size with only one-thousandth of their cost. This indicates that the pruned models have effectively inherited the parameters of the original models, enabling them to rapidly recover performance with minimal post-training and achieve convergence.

We also evaluate the post-trained models on multiple benchmarks with detailed results in  Appendix~\ref{app:sup_res_2} Table~\ref{tab:post_train_res}. The average scores for each category and the overall average are in Table~\ref{tab:avg_res_post_train}.

\begin{figure}[!tp]
    \centering
    \includegraphics[width=\linewidth,scale=0.98]{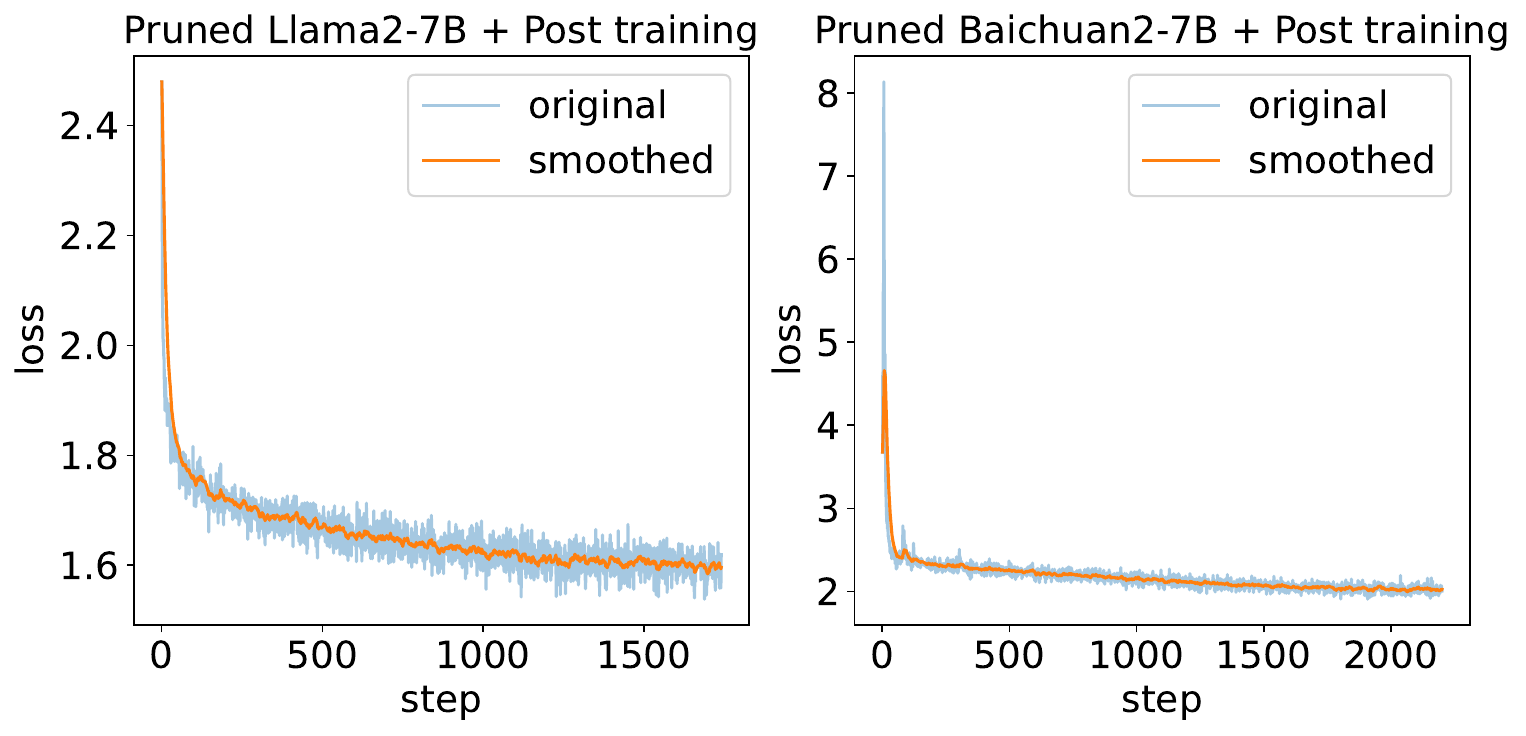}
    \caption{Loss curves for post-training.}
    \label{fig:post_train}
\end{figure}

\begin{table}[!tp]
\setlength\tabcolsep{0.6pt} 
\small
    \centering
    \begin{tabular}{c|c|c|c|c|c|c|c}
    \toprule[0.5pt]
    \midrule

    LLM & Method & Avg. & Reas. & Lan. & Know. & Exam. & Unde. \\

    \midrule
    \multirow{4}{*}{\makecell[c]{\textbf{Llama2} \\ \textbf{-7B}}} & Dense$^{*}$ & 46.55 & 60.83 & 40.67 & 68.67 & 38.89 & 33.03 \\
            \cmidrule{2-8}
            & LaCo  & 37.46 & 53.30 & 33.84 & 54.89 & 25.85 & 25.38  \\
            \cmidrule{2-8}
            & \makecell[c]{LaCo \\ +post train}  & \textbf{40.33} & \textbf{56.66} & \textbf{36.43} & \textbf{61.85} & \textbf{27.40} & \textbf{26.70} \\
            \cmidrule{2-8}
            & \makecell[c]{LaCo \\ +post train \\ +re prune}  & 32.40 & 48.07 & 20.26 & 49.46 & 25.72 & 24.56 \\
    \midrule
    \multirow{4}{*}{\makecell[c]{\textbf{Baic2.} \\ \textbf{-7B}}} & Dense$^{*}$ & 56.52 & 59.03 & 62.49 & 63.22 & 55.60 & 47.26  \\
            \cmidrule{2-8}
            & LaCo & \textbf{41.79} & 51.26 & \textbf{48.49} & 51.70 & \textbf{31.38} & 29.90 \\
            \cmidrule{2-8}
            & \makecell[c]{LaCo \\ +post train} & 40.46 & \textbf{51.67} & 40.82 & \textbf{53.97} & 27.98 & \textbf{31.28} \\
    \midrule
    \bottomrule
    \end{tabular}
    \caption{Average scores across all categories and the overall average score of pruned models, post-trained models, post-trained models followed by re-pruning.}
    \label{tab:avg_res_post_train}
\end{table}

From the tables, it is evident that the post-training of pruned Llama2-7B significantly improves its performance across various benchmarks. However, the performance of pruned Baichuan2-7B after post-training shows mixed results, with some benchmarks showing improvement while others exhibit a decrease and there is also a slight decrease in the overall score. We speculate that the pre-training data of Baichuan2-7B includes a variety of sources, resulting in a data distribution different from that of our post-training data, hindering the effectiveness of post-training. However, the consistent score improvement on pruned Llama2-7B indicates that models pruned using our LaCo indeed effectively inherit the parameters and can regain performance through low-cost post-training.

LaCo achieves excellent performance through post-training, prompting us to compare its effectiveness with the SOTA LLM-Pruner on the same training data. Our results, shown in the Appendix~\ref{app:post_train_compare} Table~\ref{tab:direct_post_train}, indicate that the model pruned by LaCo outperforms the one pruned by LLM-Pruner after post-training. Meanwhile, LaCo also significantly reduces training resource consumption.

\subsubsection{Re-pruning}
Since it is possible to partially restore performance using post-training on an LLM with approximately 25\%-30\% of its parameters pruned, it raises the question of whether we can further prune the post-trained model to obtain one with only around 50\% parameters while still maintaining relatively good performance. Thus, we further prune the previously post-trained pruned Llama2-7B model using LaCo, resulting in a model with 17 layers, retaining 55\% of the parameters of the original Llama2-7B model. We evaluate the re-pruned model. The detailed results are shown in Appendix~\ref{app:sup_res_2} Table~\ref{tab:post_train_res} and the average results are in Table~\ref{tab:avg_res_post_train}.

The tables show that even with only 55\% parameters, the model still retains about 70\% of the original 7B model performance. However, our training data quality and scale are limited. With more and better training data, LaCo should demonstrate even greater utility in the pruning+post-training+re-pruning pipeline on larger models.

\subsection{Layer-wise Similarity}
This section discusses our motivation for merging adjacent layers. Our primary motivation comes from observing that the changes in parameters and output representations between adjacent layers in the LLMs are not particularly significant.

\begin{figure}[!tp]
    \centering
    \includegraphics[width=\linewidth,scale=0.95]{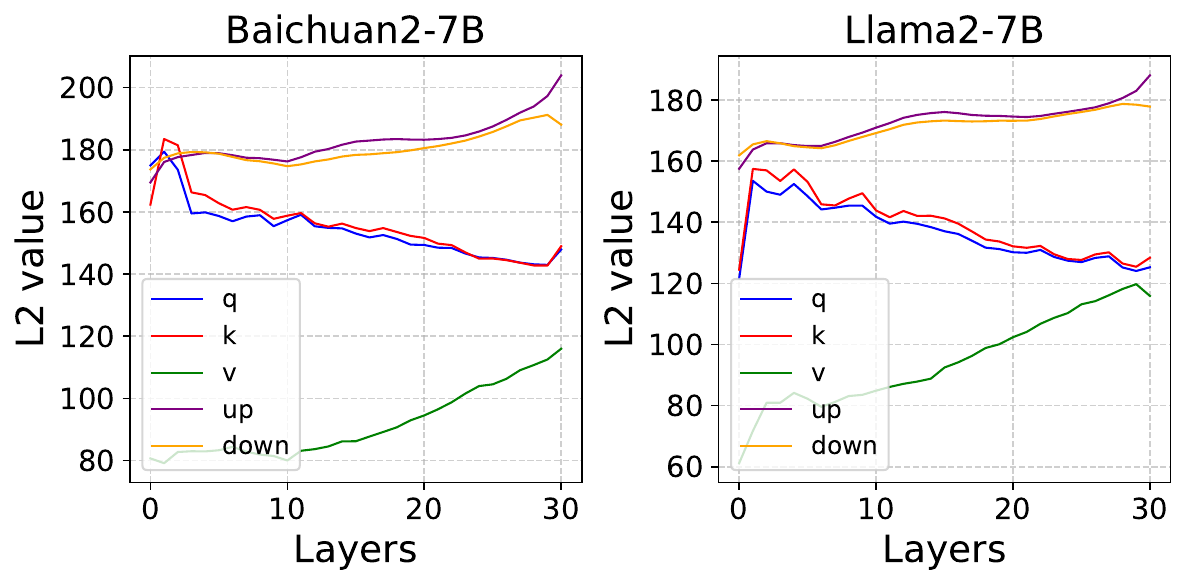}
    \caption{The L2 similarity of corresponding matrices between adjacent layers.}
    \label{fig:para_sim}
\end{figure}

In Figure~\ref{fig:para_sim}, we show the L2 similarities between the SAN q, k, v matrices of each layer and their counterparts in the subsequent layer, as well as the upscaling and downscaling matrices of the MLP for both Llama2-7B and Baichuan2-7B. The results indicate that the maximum L2 values between corresponding matrices in adjacent layers are generally no more than 200. Given the large sizes of the MLP upscaling (11008x4096) and SAN q, k, v (4096x4096) matrices, the change in each element between adjacent layers is minimal.

\begin{figure}[!tp]
    \centering
    \begin{subfigure}{0.45\linewidth}
        \centering
        \includegraphics[width=\linewidth]{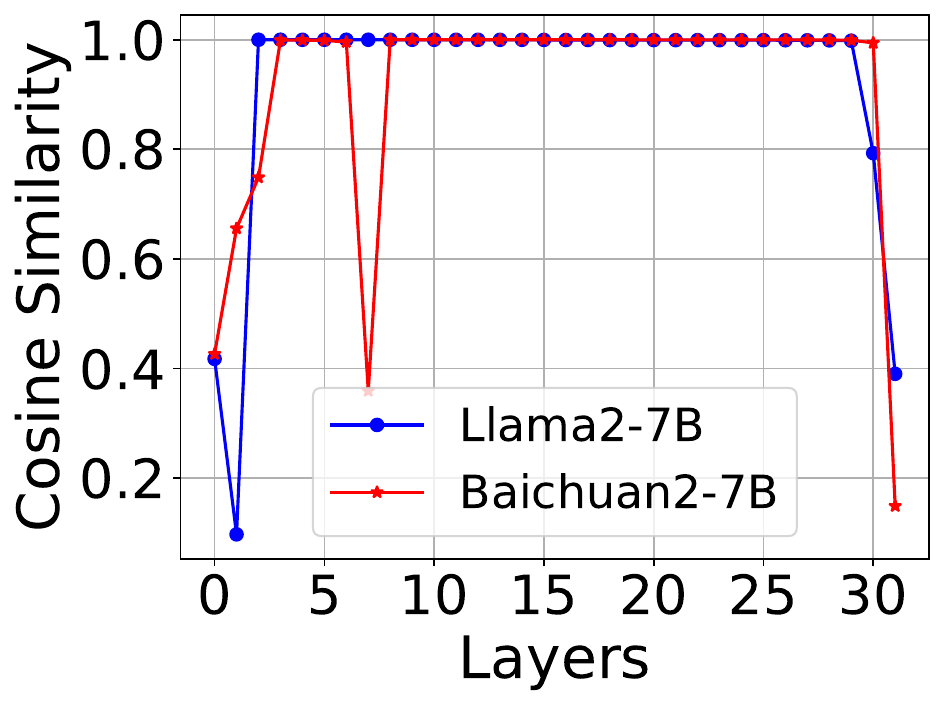}
        \caption{The cosine similarity of output representations between adjacent layers.}
        \label{fig:hid_sim}
    \end{subfigure}
    \hspace{0.01\linewidth}
    \begin{subfigure}{0.45\linewidth}
        \centering
        \includegraphics[width=\linewidth]{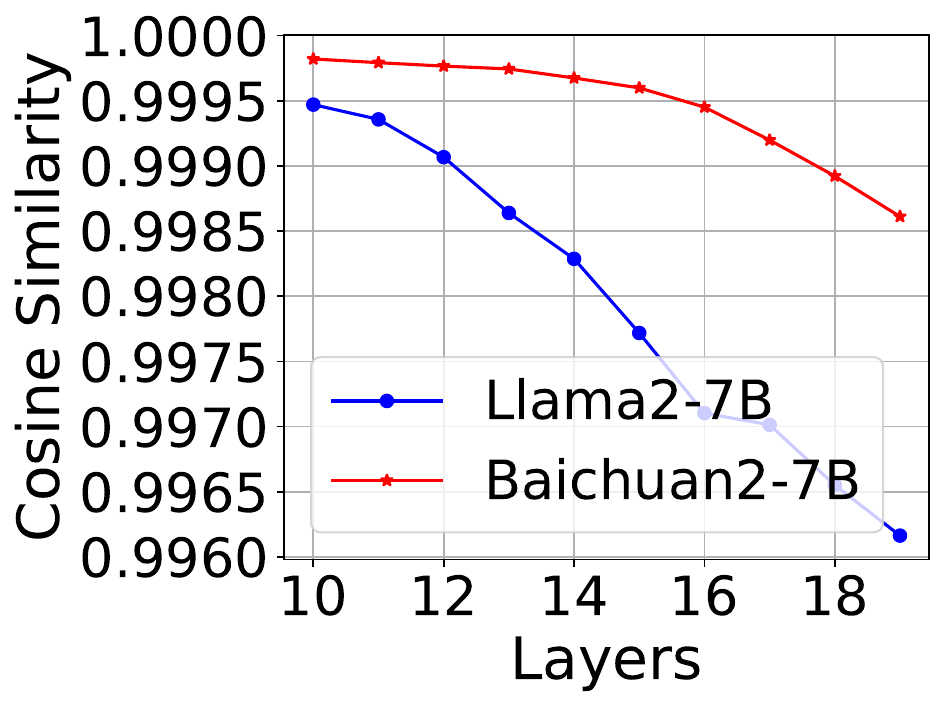}
        \caption{The similarity of output representations before and after RDSC Layer Merge.}
        \label{fig:merged_hid_sim}
    \end{subfigure}
    \caption{The cosine similarity of layer representations.}
    \label{fig:both_hid}
\end{figure}

In Figure~\ref{fig:both_hid} (a), we randomly select 20 sentences from Wikipedia and calculate the cosine similarity between the hidden-states of adjacent layers outputs. The results show that for both Baichuan2-7B and Llama2-7B, the representation similarity between adjacent layers from layers 3 to 28 is typically very close to 1. The high similarity in parameters and representations between adjacent layers leads us to consider that a single layer might replace multiple subsequent layers.

Moreover, the similarity in parameters suggests focusing on the differences between layers. Inspired by previous model merging work \cite{yu2023language,matena2022merging}, we come up with collecting parameter differences between layers and merging them into a single layer. To verify that RDSC Layer Merge can replace multiple layers with one, we conduct the experiment: we merge every four consecutive layers into one within layers 10 to 19 and evaluate the cosine similarity between the merged layer's output and the original last layer's output, as in Figure~\ref{fig:both_hid} (b), where the lowest cosine similarity on the 4096-dimensional vectors is above 0.996, confirming the effectiveness of RDSC Layer Merge in preserving representations.

\subsection{Varying Pruning Ratio}

\begin{table}[!tp]
\setlength\tabcolsep{0.6pt} 
\small
    \centering
    \begin{tabular}{c|c|c|c|c|c|c|c}
    \toprule[0.5pt]
    \midrule

    LLM & Ratio/Lay. & Avg. & Reas. & Lan. & Know. & Exam. & Unde. \\

    \midrule
    \multirow{4}{*}{\makecell[c]{\textbf{Llama2} \\ \textbf{-7B}}} & 0\%/32  & 46.55 & 60.83 & 40.67 & 68.67 & 38.89 & 33.03 \\
            \cmidrule{2-8}
            & 12.0\%/28 & 36.13 & 44.46 & \textbf{36.31} & \textbf{56.35} & \textbf{26.34} & 24.54 \\
            \cmidrule{2-8}
            & 27.1\%/23  &  \textbf{37.46} & \textbf{53.30} & 33.84 & 54.89 & 25.85 & \textbf{25.38} \\
            \cmidrule{2-8}
            & 45.0\%/17 & 30.00 & 43.66 & 19.27 & 48.06 & 24.78 & 21.44 \\
    \midrule
    \multirow{4}{*}{\makecell[c]{\textbf{Llama2} \\ \textbf{-13B}}} & 0\%/40 & 55.50 & 62.51 & 55.80 & 69.20 & 47.18 & 47.34  \\
            \cmidrule{2-8}
            & 14.6\%/34 & \textbf{53.89} & \textbf{60.56} & \textbf{54.51} & \textbf{63.58} & \textbf{46.10} & \textbf{47.46} \\
            \cmidrule{2-8}
            & 24.7\%/30 &  47.55 & 57.17 & 42.85 & 58.32 & 39.28 & 42.60 \\ 
            \cmidrule{2-8}
            & 49.7\%/20 & 38.27 & 48.20 & 26.89 & 49.26 & 32.82 & 36.58\\
    \midrule
    \bottomrule
    \end{tabular}
    \caption{Model performance at different pruning ratios.}
    \label{tab:vary_ratio}
\end{table}

In this section, we explore the performance of \textit{LaCo} at different pruning ratios. We conduct experiments on Llama2-7B and Llama2-13B, controlling the pruning ratios at approximately 10\%, 25\% (our main experiments), and around 50\% by setting different hyperparameters (as shown in Appendix~\ref{app:hyper_setting} Table~\ref{table:hparams_varying})\footnote{Further ablation study on the hyperparameters $\mathcal{C}$, $\mathcal{I}$, $\mathcal{D}$, $\mathcal{T}$, different similarity metrics, different merging strategies, different calibration datasets can be found in Appendix~\ref{app:Abla}.}. We evaluate pruned models accordingly. The average results are shown in Table~\ref{tab:vary_ratio} and the detailed results are shown in Appendix~\ref{app:sup_res_2} Table~\ref{tab:varing_res_detailed}.

As the pruning ratio increases, overall model performance decreases. However, from a pruning ratio of around 10-15\% to about 25\%, the performance does not significantly decline, indicating our method's stability within this range. Furthermore, at a pruning ratio close to 50\%, the model still maintains approximately 70\% performance, demonstrating that our method prevents model crashes even with about half of the parameters removed.

\section{Related Work}

\textbf{\textit{Model Quantization}} reduces model size by converting weights from high-precision floating points to lower-precision floating points or integers. SmoothQuant~\cite{xiao2023smoothquant} quantizes both weights and activations while smoothing activation outliers. GPTQ~\cite{frantar2022gptq} uses approximate second-order information for quantization. QLoRA~\cite{dettmers2023qlora} backpropagates gradients through a frozen, 4-bit quantized model into Low Rank Adapters. OmniQuant~\cite{shao2023omniquant} optimizes various quantization parameters.

\textbf{\textit{Knowledge Distillation}} transfers knowledge from a large model to a smaller one. Distilling step-by-step~\cite{hsieh2023distilling} trains smaller models that outperform LLMs. DISCO~\cite{chen2023disco} distills counterfactual knowledge from LLMs. SOCRATIC COT~\cite{shridhar2023distilling} distills the ability of Chain-of-Thought from LLMs. ZEPHYR~\cite{tunstall2023zephyr} applies distilled direct preference optimization to learn a chat model.

\textbf{\textit{Model Pruning}} refers to techniques for improving model efficiency by sparsification or parameter removal. Non-structured pruning often involves model sparsity. SparseGPT~\cite{frantar2023sparsegpt} reduces the pruning problem to large-scale instances of sparse regression, while SpQR~\cite{dettmers2023spqr} identifies and isolates outlier weights during LLM sparsification. Structured pruning primarily removes parts of model modules. LLM-Pruner~\cite{ma2023llm} selectively eliminates non-critical structures based on gradient information. DirectShare~\cite{cao2024head} uses activation sharing methods to improve LLM inference efficiency. ShearedLLaMA~\cite{xia2023sheared} uses targeted structured pruning and dynamic batch loading to prune Llama2. FLAP~\cite{an2024fluctuation} assesses the recoverability of the output feature map after weight removal using the fluctuation pruning metric and standardizes importance scores to adaptively define the global compressed model structure. Shortened LLaMA~\cite{kim2024shortened} demonstrates that depth pruning can efficiently compress LLMs while matching or surpassing the performance of recent width pruning methods.

However, model quantization and sparsification typically require special hardware and usually impact performance. Knowledge distillation is costly and task-specific. Existing structured pruning methods often disrupt the model inherent structure. In contrast, \textit{LaCo} maintains the model structure, which is more concise and preserves excellent performance. Although some existing works~\cite{din2023jump, fan2019reducing, belrose2023eliciting} have utilized layer-skipping/dropping to accelerate inference, LaCo is fundamentally different. It is the first pruner based on layer collapse, resulting in a smaller, faster, more memory-efficient model with strong performance. Furthermore, those methods typically require training new parameters to determine which layers to skip/drop during inference, whereas LaCo does not require any training.

\section{Conclusion}
In this paper, we propose a concise layer-wise structured pruning method called \textit{Layer Collapse (LaCo)}, which merges rear model layers into preceding layers for rapid model size reduction. \textit{LaCo} does not require special hardware support and preserves the model's intrinsic structure. Experimental results show that \textit{LaCo} significantly outperforms current SOTA structured pruning methods, also revealing potential parameter redundancy in existing LLMs. We conduct ablation studies on various settings of \textit{LaCo}. We also post-train the pruned models, confirming that \textit{LaCo} effectively inherits the original model parameters. Additionally, we discuss our motivation from the perspective of layer-wise similarity and explore the performance of \textit{LaCo}-pruned models at different pruning ratios.

\section*{Limitations}

Due to \textit{LaCo}'s pruning process primarily relying on layer-wise iterations, it cannot directly control the pruning ratio like previous methods. Instead, it requires tuning hyperparameters such as the representation similarity threshold $\mathcal{T}$ for control. In future work, we will summarize additional experimental patterns regarding how to set hyperparameters to achieve a specific pruning ratio.

Our motivation comes from current model merging techniques, but like existing baselines (LLM-Pruner~\cite{ma2023llm} and SliceGPT~\cite{ashkboos2024slicegpt}), our method lacks a complete theoretical proof. We consider this as future work.

Additionally, there may be better merging methods, even though our experimental results demonstrate that LaCo's current merging approach is effective. We will continue to search for improved layer merging methods in the future.

\bibliography{anthology,reference_new}
\bibliographystyle{acl_natbib}

\clearpage
\appendix

\section{Hyperparameter Settings} \label{app:hyper_setting}

\begin{table}[!htp]
\setlength\tabcolsep{0.8pt} 
\centering
\tabcolsep=0.3cm
{\resizebox{0.90\linewidth}{!}{
\begin{tabular}{l|ccccc}
\toprule
\hline
LLM & \textbf{$\mathcal{C}$} & \textbf{$\mathcal{L}$} & \textbf{$\mathcal{H}$} & \textbf{$\mathcal{I}$} & \textbf{$\mathcal{T}$} \\ \midrule
\begin{tabular}[c]{@{}l@{}}\textbf{Llama2-7B} \end{tabular}  & 4 & 1 & 32  & 2 & 0.65 \\ 
\midrule
\begin{tabular}[c]{@{}l@{}}\textbf{Llama2-13B} \end{tabular}  & 6 & 1 & 40 & 2 & 0.75 \\ 
\midrule
\begin{tabular}[c]{@{}l@{}}\textbf{Baichuan2-7B} \end{tabular}  & 4 & 1 & 32 & 2 & 0.70 \\ 
\midrule
\begin{tabular}[c]{@{}l@{}}\textbf{Llama2-13B} \end{tabular}  & 6 & 1 & 40 & 2 & 0.70 \\ 
\hline
\bottomrule
\end{tabular}}}
\caption{Hyperparameter settings for main results.}
\label{table:hparams}
\end{table}

\begin{table}[!htp]
\setlength\tabcolsep{0.8pt} 
\centering
\tabcolsep=0.2cm
{\resizebox{0.90\linewidth}{!}{
\begin{tabular}{l|ccccc}
\toprule
\hline
LLM (Ratio/Lay.) & \textbf{$\mathcal{C}$} & \textbf{$\mathcal{L}$} & \textbf{$\mathcal{H}$} & \textbf{$\mathcal{I}$} & \textbf{$\mathcal{T}$} \\ \midrule
\begin{tabular}[c]{@{}l@{}}\textbf{Llama2-7B (12.0\%/28)} \end{tabular}  & 5 & 1 & 32  & 2 &  0.85 \\ 
\begin{tabular}[c]{@{}l@{}}\textbf{Llama2-7B (27.1\%/23)} \end{tabular}  & 4 & 1 & 32  & 2 & 0.65 \\ 
\begin{tabular}[c]{@{}l@{}}\textbf{Llama2-7B (45.0\%/17)} \end{tabular}  & 6 & 1 & 32  & 2 & 0.45 \\ 
\midrule
\midrule
\begin{tabular}[c]{@{}l@{}}\textbf{Llama2-13B (14.6\%/34)} \end{tabular}  & 7 & 1 & 40  & 2 & 0.85 \\ 
\begin{tabular}[c]{@{}l@{}}\textbf{Llama2-13B (24.7\%/30)} \end{tabular}  & 6 & 1 & 40  & 2 & 0.75 \\ 
\begin{tabular}[c]{@{}l@{}}\textbf{Llama2-13B (49.7\%/20)} \end{tabular}  & 7 & 1 & 40  & 2 & 0.45 \\ 
\hline
\bottomrule
\end{tabular}}}
\caption{Hyperparameters for varying pruning ratios.}
\label{table:hparams_varying}
\end{table}

\section{Ablation Study}\label{app:Abla}
\subsection{Impact of Number of Layers Merged per Operation $\mathcal{C}$}
We conduct an ablation study on the number of layers to be merged during each merging operation $\mathcal{C}$, as it is one of the key parameters that control the compression rate. The results are shown in Table~\ref{tab:varying_c}, where the experiment is based on the Llama2-7b with a 27\% compression rate.

\begin{table}[!htbp]
\small
    \centering
    \resizebox{0.45\textwidth}{!}{
    \begin{tabular}{c|c|c|c}
    \toprule[0.5pt]
    \midrule
          & BoolQ & PIQA & HeSw \\
    \midrule
    \textbf{$\mathcal{C}=1$ (Result in Table~\ref{tab:merge_or_drop})} & 52.57	& 68.17 &	48.61 \\
    \midrule
    \textbf{$\mathcal{C}=2$} & 63.12 & 68.93 & 	54.98 \\
    \midrule
    \textbf{$\mathcal{C}=3$} & 63.98 & 69.71 & 	55.34 \\
    \midrule
    \textbf{$\mathcal{C}=4$ (Result in Table~\ref{tab:main_res})} & 64.07 & 69.80 & 55.69 \\
    \midrule
    \textbf{$\mathcal{C}=5$} & 64.15 & 70.02 &	55.76 \\
    \midrule
    \textbf{$\mathcal{C}=6$} & 64.00 & 69.89 & 	55.49 \\
    \midrule
    \bottomrule
    \end{tabular}
    }
    \caption{Ablation study on $\mathcal{C}$.}
    \label{tab:varying_c}
\end{table}

When $\mathcal{C}$ is too small, the model's performance degrades. When $\mathcal{C}$ is between 3 and 6, the performance remains relatively good. If $\mathcal{C}$ is too large, it causes the cosine similarity of the hidden-states to drop quickly below the threshold, stopping the pruning loop prematurely and resulting in a lower compression rate. Therefore, we recommend setting $\mathcal{C}$ between 4 and 7.

\subsection{Impact of Minimum Interval $\mathcal{I}$}

To explore the impact of different values of $\mathcal{I}$, we also conduct an ablation study. The experiment is based on the Llama2-7b model with a 27\% sparsity. The results are shown in Table~\ref{tab:varying_I}.

\begin{table}[!htbp]
\small
    \centering
    \resizebox{0.45\textwidth}{!}{
    \begin{tabular}{c|c|c|c}
    \toprule[0.5pt]
    \midrule
          & BoolQ & PIQA & HeSw \\
    \midrule
    \textbf{$\mathcal{I}=1$} & 63.99 & 68.90 & 55.79 \\
    \midrule
    \textbf{$\mathcal{I}=2$ (Result in Table~\ref{tab:main_res})} & 64.07 & 69.80 & 55.69 \\
    \midrule
    \textbf{$\mathcal{I}=3$} & 64.05 & 69.57	& 55.48 \\
    \midrule
    \textbf{$\mathcal{I}=4$} & 62.29 & 68.92 & 54.87 \\
    \midrule
    \bottomrule
    \end{tabular}
    }
    \caption{Ablation study on $\mathcal{I}$.}
    \label{tab:varying_I}
\end{table}

The results indicate that the model performs well when $\mathcal{I}$ is between 1 and 3. However, when $\mathcal{I}$ is set to 4, the model's performance declines. This observation aligns with our empirical findings: modifying the parameters of layers closer to the input end (such as layers 1-12) can lead to performance drops. Setting $\mathcal{I}$ to 4 causes the merge operations to occur closer to the input end. Therefore, we choose $\mathcal{I} = 2$, as it relatively maintains better model performance.

\subsection{Impact of Dataset $\mathcal{D}$}
To understand the impact of different datasets $\mathcal{D}$ on LaCo's effectiveness, we conduct an ablation study using Llama2-7B. We perform three rounds of pruning, each time selecting different sets of 10 sentences as $\mathcal{D}$, with a 27\% compression ratio to match Table~\ref{tab:main_res}. The results are shown in Table~\ref{tab:abla_D}.

\begin{table}[!htbp]
\small
    \centering
    \begin{tabular}{c|c|c|c}
    \toprule[0.5pt]
    \midrule
          & BoolQ & PIQA & HeSw \\
    \midrule
    \textbf{Result in Table~\ref{tab:main_res}} & 64.07 & 69.80 & 55.69 \\
    \midrule
    \textbf{Round1} & 64.50 & 70.28 & 56.14 \\
    \textbf{Round2} & 63.70 & 69.14 & 54.92 \\
    \textbf{Round3} & 63.90 & 69.66 & 55.18 \\
    \midrule
    \bottomrule
    \end{tabular}
    \caption{Ablation study on $\mathcal{D}$, where the model is Llama2-7B with compression ratio of 27\%.}
    \label{tab:abla_D}
\end{table}

We can find that the results are all on nearly the same scale, indicating that the random selection of sentences in $\mathcal{D}$ has no notable impact on the results.

\subsection{Impact of Threshold $\mathcal{T}$}
We also aim to understand the impact of threshold $\mathcal{T}$ on the model pruned by LaCo. For the results of Llama2-7B in Table~\ref{tab:main_res}, we keep other parameters unchanged and set $\mathcal{T}$ to 0.85, 0.45, and 0.25 ($\mathcal{T}=0.65$ corresponds to the results in Table~\ref{tab:main_res}). We evaluate the pruned models on several datasets. The results, shown in Table~\ref{tab:abla_T}, indicate that a smaller threshold leads to a larger compression ratio and poorer performance, which aligns with intuition.

\begin{table}[!htbp]
\small
    \centering
    \resizebox{0.45\textwidth}{!}{
    \begin{tabular}{c|c|c|c|c}
    \toprule[0.5pt]
    \midrule
          & Ratio/Layer & BoolQ & PIQA & HeSw \\
    \midrule
    \textbf{$\mathcal{T}=0.65$} & 27.1\%/23 & 64.07 & 69.80 & 55.69 \\
    \midrule
    \textbf{$\mathcal{T}=0.85$} & 9.0\%/29 & 70.92 & 76.01 & 68.15 \\
    \textbf{$\mathcal{T}=0.45$} & 48.0\%/16 & 59.82 & 60.34 & 36.09 \\
    \textbf{$\mathcal{T}=0.25$} & 60.1\%/12 & 41.68 & 51.74 & 26.37 \\
    \midrule
    \bottomrule
    \end{tabular}
    }
    \caption{Ablation study on $\mathcal{T}$. The model is Llama2-7B. $\mathcal{T}=0.65$ corresponds to the results in Table~\ref{tab:main_res}.}
    \label{tab:abla_T}
\end{table}

\subsection{Different Similarity Metrics}
we use the cosine similarity of the representations from the final layer outputs as a metric for LaCo. We also aim to explore the feasibility of using common distribution distances as metrics, such as KL divergence and kernel/linear CKA (Centered Kernel Alignment~\cite{kornblith2019similarity}). Specifically, we replace cosine similarity with distribution distances to measure the difference in representations. However, we find that the distribution distances almost remain constant, as in Table~\ref{tab:abla_metric}.

\begin{table}[!htbp]
\centering
\resizebox{0.45\textwidth}{!}{
    \begin{tabular}{c|c|c|c}
    \toprule[0.5pt]
    \midrule
    & KL Divergence & Kernel CKA & Linear CKA  \\
    \midrule
    \textbf{Constant} & 0.00 & 1.00 & 1.00 \\
    \midrule
    \bottomrule
    \end{tabular}
}
\caption{The distribution distances of model output representations.}
\label{tab:abla_metric}
\end{table}

KL Divergence being nearly 0 and CKA being nearly 1 indicate minimal differences in the distributions of the final layer outputs. This suggests that the LaCo merging process does not significantly alter the model output. Additionally, due to the high dimensionality of the representations, the distribution distances tend to be constant, resulting in a lack of discrimination. Therefore, we chose the more discriminative cosine similarity.

\subsection{Drop or Merge}
As shown in Eq.~\ref{eq:merge}, LaCo merges multiple adjacent layers into one. This leads us to consider an extreme case: if we set $m=1$, LaCo will no longer merge layers but simply drop a layer. We also aim to explore the performance differences between the drop and merge operations in LaCo. Thus, we conduct an experiment with $m=1$ on Llama2-7B, setting the compression rate to the same 27\% as in Table~\ref{tab:main_res}. The results on some benchmarks are shown in Table~\ref{tab:merge_or_drop}.

\begin{table}[!htbp]
\small
    \centering
    \resizebox{0.45\textwidth}{!}{
    \begin{tabular}{c|c|c|c}
    \toprule[0.5pt]
    \midrule
          & BoolQ & PIQA & HeSw \\
    \midrule
    \textbf{Drop} & 52.57 & 68.17 & 48.61 \\
    \midrule
    \textbf{Mege (Result in Table~\ref{tab:main_res})} & 64.07 & 69.80 & 55.69 \\
    \midrule
    \bottomrule
    \end{tabular}
    }
    \caption{Results of the drop or merge operation.}
    \label{tab:merge_or_drop}
\end{table}

The results indicate that the drop operation is not as effective as the merge operation and $m=1$ is not a good hyperparameter setting. The merge operation aligns better with our intention.

\subsection{Iterative-based or Rule-based Merge}
\begin{table}[!htp]
\setlength\tabcolsep{1.5pt} 
\small
    \centering
    \begin{tabular}{c|c|c|c|c}
    \toprule[0.5pt]
    \midrule
    LLM & Strategy & BoolQ & PIQA & HeSw \\
    \midrule
    \multirow{2}{*}{\textbf{Llama2-7B}} & Rule & 63.49 & 68.72 & 53.16  \\
            \cmidrule{2-5}
            & LaCo & \textbf{64.07} & \textbf{69.80} & \textbf{55.69}\\
    \midrule
    \multirow{2}{*}{\textbf{Baichuan2-7B}} & Rule & 52.94	& 67.28 & 48.78  \\
    \cmidrule{2-5}
    & LaCo & \textbf{56.15} & \textbf{68.50} & \textbf{52.28}\\
    \midrule
    \bottomrule
    \end{tabular}
    \caption{The results of using rule-based merging and LaCo iterative merging.}
    \label{tab:laco_or_rule}
\end{table}
We want to determine if our iterative search-based merging strategy is superior to rule-based merging. To test this, we perform rule-based merging on Llama2-7B and Baichuan2-7B, both with 32 layers. We merge layers in groups of four, starting from the top, specifically merging layers (29, 30, 31, 32), (21, 22, 23, 24), and (13, 14, 15, 16). We avoid merging before the 16th layer due to significant performance drops observed in those cases. The resulting models achieved compression rates equivalent to those in Table~\ref{tab:main_res}. The results in Table~\ref{tab:laco_or_rule} indicate that LaCo performs better than the rule-based approach. Notably, even simple rule-based merging can outperform baselines across multiple datasets, demonstrating the potential of merging for model compression.

\subsection{Effect of Calibration Dataset from Different Sources}

We explore whether calibration datasets from different sources will affect the performance.

Specifically, we select 10 sentences from the BookCorpus~\cite{soskkobayashi2018bookcorpus} dataset used by LLM-Pruner as the calibration dataset. We prune Llama2-7B at a 27\% compression rate and evaluate the results on several datasets. We repeat the experiments three times.

\begin{table}[!htbp]
\small
    \centering
    \resizebox{0.45\textwidth}{!}{
    \begin{tabular}{c|c|c|c}
    \toprule[0.5pt]
    \midrule
          & BoolQ & PIQA & HeSw \\
    \midrule
    \textbf{Wikipedia} & 64.07 & 69.80 & 55.69 \\
    \midrule
    \textbf{BookCorpus-Round1} & 64.15 & 69.60 & 55.87 \\
    \midrule
    \textbf{BookCorpus-Round2} & 63.80 &	69.45 &	55.61 \\
    \midrule
    \textbf{BookCorpus-Round3} & 64.30 & 70.08 & 55.98 \\
    \midrule
    \bottomrule
    \end{tabular}
    }
    \caption{Results of calibration datasets from different sources.}
    \label{tab:dif_cal}
\end{table}

The results in Table~\ref{tab:dif_cal} are at the same level as those in our paper. This demonstrates that, given high-quality corpora, different calibration datasets have little impact on the performance of the models pruned with LaCo at the same compression rate.

\subsection{Effect of Calibration Dataset Size}
We want to explore how the size of the calibration dataset affects the performance of LaCo. Specifically, we take 5, 10, 50, 100, and 200 samples from Wikipedia and prune Llama2-7B at a 27\% compression rate. The results are shown in the Table~\ref{tab:dif_size}, and it can be seen that the size of the calibration dataset has little impact on the results.

\begin{table}[!htbp]
\small
    \centering
    \resizebox{0.45\textwidth}{!}{
    \begin{tabular}{c|c|c|c}
    \toprule[0.5pt]
    \midrule
          & BoolQ & PIQA & HeSw \\
    \midrule
    \textbf{Result in Table~\ref{tab:main_res} (10 samples)} & 64.07 & 69.80 & 55.69 \\
    \midrule
    \textbf{5 samples} & 63.89 & 69.74 & 55.54 \\
    \midrule
    \textbf{50 samples} & 64.10	& 69.69	& 55.81 \\
    \midrule
    \textbf{100 samples} & 64.25 & 69.81 & 55.59\\
    \midrule
    \textbf{200 samples} & 63.98 & 69.92 & 55.62 \\
    \midrule
    \bottomrule
    \end{tabular}
    }
    \caption{Results of different calibration dataset size.}
    \label{tab:dif_size}
\end{table}

Therefore, we chose a small calibration dataset to ensure both the pruning effect and the pruning speed. The size of our calibration dataset is the same as that used by the LLM-Pruner, which is 10 samples.

\section{Post-Training Implementation Details} \label{app:impl_det}
We use the LLaMA-Factory~\cite{zheng2024llamafactory} framework along with DeepSpeed ZeRO-2. The sequence length is set to 4096, following the default settings for Llama2-7B and Baichuan2-7B. We use the Adam optimizer with a learning rate of 2e-4, setting $\beta_1=0.9$ and $\beta_2=0.95$. The batch size is 8 per GPU, resulting in a total batch size of 32, with gradient accumulation steps set to 4. We employ a cosine learning rate scheduler, apply a weight decay of 0.1, and set the maximum gradient normalization to 1.0.

\section{Supplementary Results (Part 1)} \label{app:sup_res}

\subsection{Results on Llama2-70B} \label{app:exp_llama_70}
We supplement the results using Llama2-70B with a compression rate set to 30\%. We conduct experiments on CMNLI, HeSw, PIQA, and BoolQ, and Table~\ref{tab:exp_70} presents the average results.

\begin{table}[!htbp]
\small
    \centering
    \begin{tabular}{c|c|c|c|c}
    \toprule[0.5pt]
    \midrule
     Pruner & Dense & LaCo & LLM-Pruner & SliceGPT \\
    \midrule
    Average & 65.92 & 57.91 & 53.78 & 51.36 \\
    \midrule
    \bottomrule
    \end{tabular}
    \caption{The average scores on several benchmarks using different pruners with a 30\% compression rate on Llama2-70B.}
    \label{tab:exp_70}
\end{table}

The results demonstrate that our method can also scale up effectively. It performs better than the baselines even on the 70B model.

\subsection{Memory Consumption and Inference Speed} \label{app:full_mem_infer}
\begin{table}[!htp]
\setlength\tabcolsep{0.6pt} 
\small
    \centering
    \begin{tabular}{c|c|c|c}
    \toprule[0.5pt]
    \midrule

    LLM & Pruner & Memory (MB) & Infer. (tokens/s) \\

    \midrule
    \multirow{4}{*}{\makecell[c]{\textbf{Llama2} \\ \textbf{-7B}}} & Dense  & 13410 & 38.53  \\
            \cmidrule{2-4}
            & LLMPru. & 10434 & 33.22 ($\downarrow$) \\
            \cmidrule{2-4}
            & SliceGPT & 11770 & 44.88 \\
            \cmidrule{2-4}
            & LaCo & \textbf{9894} & \textbf{50.80}  \\
    \midrule
    \multirow{4}{*}{\makecell[c]{\textbf{Llama2} \\ \textbf{-13B}}} & Dense  & 25902 & 29.98  \\
            \cmidrule{2-4}
            & LLMPru. & 19874 & 27.15 ($\downarrow$) \\
            \cmidrule{2-4}
            & SliceGPT & 22506 & 35.16 \\
            \cmidrule{2-4}
            & LaCo & \textbf{19422} & \textbf{38.65}  \\
    \midrule
    \multirow{4}{*}{\makecell[c]{\textbf{Baic2.} \\ \textbf{-7B}}} & Dense  & 14810 & 37.13  \\
            \cmidrule{2-4}
            & LLMPru. & 11898 & 38.95 ($\downarrow$) \\
            \cmidrule{2-4}
            & SliceGPT & 13586 & 36.67 \\
            \cmidrule{2-4}
            & LaCo & \textbf{11716} & \textbf{49.15}  \\
    \midrule
    \multirow{4}{*}{\makecell[c]{\textbf{Baic2.} \\ \textbf{-13B}}} & Dense  & 27410 & 36.93  \\
            \cmidrule{2-4}
            & LLMPru. & 22390 & 31.61 ($\downarrow$) \\
            \cmidrule{2-4}
            & SliceGPT & 23956 & 29.35 ($\downarrow$) \\
            \cmidrule{2-4}
            & LaCo & \textbf{21010} & \textbf{47.46}  \\
    \midrule
    \bottomrule
    \end{tabular}
    \caption{The memory consumption and average inference speed on English Wikipedia dataset for different pruned models. $\downarrow$ means the performance worse than the Dense model.}
    \label{tab:all_mem_infer}
\end{table}

\subsection{Comparison of Post-trained Pruned Models} \label{app:post_train_compare}
\begin{table}[!htp]
\setlength\tabcolsep{2.0pt} 
\small
    \centering
    \begin{tabular}{c|c|c|c|c}
    \toprule[0.5pt]
    \midrule
     & GPU*hour  & BoolQ & PIQA & HeSw \\
    \midrule
    \textbf{LaCo} & \textbf{88} & \textbf{60.26} & \textbf{65.01} & \textbf{45.49}  \\
    \midrule
    \textbf{LLM-Pru.} &  216 & 58.75 &	61.26 & 43.53  \\
    \midrule
    \bottomrule
    \end{tabular}
    \caption{Results on different datasets for models pruned to 55\% sparsity using LaCo and LLM-Pruner on Llama2-7B, followed by post-training on the same data.}
    \label{tab:direct_post_train}
\end{table}
We prune Llama2-7B to 55\% sparsity using LaCo and LLM-Pruner, then conduct post-training with the same data in Table~\ref{tab:avg_res_post_train}. The results in Table~\ref{tab:direct_post_train} show that LaCo performs better. Additionally, training the model pruned by LLM-Pruner requires 216 GPU*hours (27 hours on 8 A100 GPUs), while only 88 GPU*hours (22 hours on 4 A100 GPUs) for LaCo-pruned model. Thus, LaCo saves more computational resources for post-training.

\subsection{Performance Percentage for Main Results} \label{app:perf_prec}
\clearpage
\begin{table*}[!htp]
\setlength\tabcolsep{0.8pt} 
\small
    \centering
    \resizebox{0.97\linewidth}{!}{
    \begin{tabular}{c|c|c|ccc|ccc|cc|cc|cccc}
    \toprule[0.5pt]
    \midrule
    \multirow{2}{*}{LLM} & \multirow{2}{*}{Pruner} & \multirow{2}{*}{Ratio/Lay.} & \multicolumn{3}{c|}{Reasoning(\%)} & \multicolumn{3}{c|}{Language(\%)} & \multicolumn{2}{c|}{Knowledge(\%)} & \multicolumn{2}{c|}{Examination(\%)} & \multicolumn{4}{c}{Understanding(\%)} \\
    & &  & CMNLI & HeSw & PIQA & CHID & WSC$_{\text{P}}$ & WSC$_{\text{G}}$ & CSQA & BoolQ & MMLU & CMMLU &  Race$_{\text{H}}$ &  Race$_{\text{M}}$ & XSum & C3 \\
    \midrule
    \multirow{4}{*}{\makecell[c]{\textbf{Llama2} \\ \textbf{-7B}}} & Dense$^{*}$ & 0\%/32 & 100 & 100 & 100 & 100 & 100 & 100 & 100 & 100 & 100 & 100 & 100 & 100 & 100 & 100  \\
            \cmidrule{2-17}
            & LLMPru.  & 27.0\%/32 & 104.09 & \textbf{79.13} & \textbf{91.10} & 54.84 & 97.44 & 2.50 & 63.76 & 78.11 & 50.81 & 79.25 & 63.53 & 67.42 & 58.49 & 58.57 \\
            \cmidrule{2-17}
            & SliceGPT  & 26.4\%/32 & 96.12 & 70.46 & 84.69 & 45.16 & 97.44 & 50.00 & 62.04 & 54.22 & \textbf{62.98} & \textbf{79.63} & 59.34 & 65.34 & 24.85 & \textbf{90.86}\\
            \cmidrule{2-17}
            & \textbf{LaCo}  & \textbf{27.1\%/23} & \textbf{104.40} & 78.05 & 89.28 & \textbf{78.50} & \textbf{107.68} & \textbf{65.00} & \textbf{68.55} & \textbf{90.66} & 57.60 & 79.22 & \textbf{63.67} & \textbf{71.22} & \textbf{79.47} & 90.61 \\
    \midrule
    \multirow{4}{*}{\makecell[c]{\textbf{Llama2} \\ \textbf{-13B}}} & Dense$^{*}$  & 0\%/40 & 100 & 100 & 100 & 100 & 100 & 100 & 100 & 100 & 100 & 100 & 100 & 100 & 100 & 100 \\
            \cmidrule{2-17}
            & LLMPru.  & 24.4\%/40 & \textbf{100.12} & \textbf{90.55} & \textbf{96.17} & 67.28 & 79.24 & 0.00 & 76.01 & 78.91 & 45.32 & 63.78 & 38.72 & 36.65 & \textbf{81.37} & 68.05\\
            \cmidrule{2-17}
            & SliceGPT  & 23.6\%/40 & 90.39 & 74.45 & 86.61 & 36.45 & 71.70 & \textbf{57.58} & 70.63 & 52.95 & 66.76 & 66.57 & 40.34 & 39.89 & 22.37 & 88.23 \\
            \cmidrule{2-17}
            & \textbf{LaCo}  & \textbf{24.6\%/30} & 99.61 & 86.05 & 93.18 & \textbf{75.70} & \textbf{103.77} & 56.07 & \textbf{78.70} & \textbf{89.48} & \textbf{82.56} & \textbf{84.20} & \textbf{93.90} & \textbf{93.87} & 61.33 & \textbf{94.57} \\
    \midrule
    \multirow{4}{*}{\makecell[c]{\textbf{Baic2.} \\ \textbf{-7B}}} & Dense$^{*}$  & 0\%/32 & 100 & 100 & 100 & 100 & 100 & 100 & 100 & 100 & 100 & 100 & 100 & 100 & 100 & 100 \\
            \cmidrule{2-17}
            & LLMPru. & 24.2\%/32 & 96.73 & \textbf{79.43} & \textbf{94.29} & 84.43 & \textbf{130.23} & 0.00 & \textbf{75.75} & \textbf{96.67} & 45.95 & 45.11 & 41.73 & 43.65 & \textbf{76.68} & 64.51 \\
            \cmidrule{2-17}
            & SliceGPT  & 22.2\%/32 & 96.10 & 37.43 & 66.08 & 17.96 & 88.37 & 0.00 & 30.99 & 62.09 & 46.41 & 44.34 & 44.71 & 44.06 & 0.00 & 41.18 \\
            \cmidrule{2-17}
            & \textbf{LaCo} & \textbf{24.2\%/23} & \textbf{98.89} & 77.38 & 89.93 & \textbf{92.22} & 102.32 & \textbf{42.42} & 74.85 & 88.70 & \textbf{58.12} & \textbf{54.86} & \textbf{55.08} & \textbf{54.31} & 57.73 & \textbf{78.78} \\
    \midrule
    \multirow{4}{*}{\makecell[c]{\textbf{Baic2.} \\ \textbf{-13B}}} & Dense$^{*}$  & 0\%/40 & 100 & 100 & 100 & 100 & 100 & 100 & 100 & 100 & 100 & 100 & 100 & 100 & 100 & 100 \\
            \cmidrule{2-17}
            & LLMPru. & 24.3\%/40 & \textbf{101.78} & 75.34 & \textbf{91.99} & 87.50 & 90.69 & 0.00 & 59.18 & 84.39 & 39.43 & 41.10 & 31.47 & 31.35 & \textbf{54.79} & 60.77 \\
            \cmidrule{2-17}
            & SliceGPT  & 22.8\%/40 & 96.57 & 36.36 & 65.36 & 12.50 & 88.37 & 0.00 & 27.47 & 56.46 & 39.02 & 41.23 & 32.05 & 31.22 & 0.00 & 38.07 \\
            \cmidrule{2-17}
            & \textbf{LaCo}  & \textbf{24.7\%/30} & 99.46 & \textbf{85.39} & 88.23 & \textbf{92.26} & \textbf{106.96} & \textbf{95.46} & \textbf{84.53} & \textbf{93.06} & \textbf{87.32} & \textbf{87.56} & \textbf{84.61} & \textbf{83.84} & 49.38 & \textbf{93.08} \\
    \midrule
    \bottomrule
    \end{tabular}}
    \caption{The percentage of each model's score on each benchmark relative to the score of Dense$^{*}$ in the main results. Models pruned by LaCo maintain performance above 70\% on most benchmarks and avoid crashes, with no performance falling below 30\%.}
    \label{tab:main_res_per}
\end{table*}

\subsection{Examples of Responses} \label{app:ex_res}
\begin{table}[!htbp]
    \centering
    \resizebox{0.95\linewidth}{!}{
    \Large
    \begin{tabular}{p{3cm}|p{10cm}}
    \toprule
    \midrule
    \textbf{Prompt} & Document: The 18-year-old scored 88.40 to make history in what was the fifth and the final stop of the World Cup season.\textbackslash nShe came ahead of Sweden's Emma Dahlstrom and Swiss Mathilde Gremaud.\textbackslash nBoston-born Atkin, who initially competed for the US before switching to Great Britain aged 15, was making her 15th appearance at a World Cup event.\textbackslash nAtkin will be competing at the Freestyle World Championships in Sierra Nevada, Spain (9-19 March). The event will be live on the BBC Sport website, app, connected TV and red button.\textbackslash nBased on the previous text, provide a brief single summary: \\
    \midrule
    \midrule
    \textbf{Pruner} &   \textbf{Generated Responses} \\
    \midrule
    \textbf{LLMPru.} & \textbackslash n\textbackslash n\textbackslash n\textbackslash n\textbackslash n\textbackslash n\textbackslash n\textbackslash n\textbackslash n\textbackslash n\textbackslash n\textbackslash n\textbackslash n\textbackslash n\textbackslash n\textbackslash n\textbackslash n\textbackslash n\textbackslash n \\
    \midrule
    \textbf{SliceGPT} & of the 19900s of the 1900s of the 1900s of the 1900s. \\
    \midrule
    \textbf{LaCo} & Boston-born Atkin, who initially competed for the US before switching to Britain aged 15, was making her 15th appearance at a World Cup event.\textbackslash nThe 18-year-old scored 88.40 to make history in what was the fifth and the final stop of the World Cup season. \\
  \midrule
  \bottomrule
    \end{tabular}
    }
    \caption{A response on the Xsum benchmark from Llama2-7B after pruning with different pruners. In this case, the models pruned by the baseline pruners generate repetitive and meaningless text, while only LaCo is able to smoothly respond with meaningful text according to the instructions.}
    \label{tab:response_dif_pruner}
\end{table}

\section{Supplementary Results (Part 2)} \label{app:sup_res_2}
\begin{table*}[!ht]
\setlength\tabcolsep{0.8pt} 
    \centering
    \resizebox{0.98\linewidth}{!}{
    \small
    \begin{tabular}{c|c|ccc|ccc|cc|cc|cccc}
    \toprule[0.5pt]
    \midrule
    \multirow{2}{*}{LLM} & \multirow{2}{*}{Method} & \multicolumn{3}{c|}{Reasoning} & \multicolumn{3}{c|}{Language} & \multicolumn{2}{c|}{Knowledge} & \multicolumn{2}{c|}{Examination} & \multicolumn{4}{c}{Understanding} \\
    & & CMNLI & HeSw & PIQA & CHID & WSC$_{\text{P}}$ & WSC$_{\text{G}}$ & CSQA & BoolQ & MMLU & CMMLU &  Race$_{\text{H}}$ &  Race$_{\text{M}}$ & XSum & C3 \\
    
    \midrule
    \multirow{4}{*}{\makecell[c]{\textbf{Llama2} \\ \textbf{-7B}}} & Dense$^{*}$ & 32.98 & 71.35 &  78.18 & 46.04 & 37.50 & 38.46 & 66.67 & 70.67 & 45.92 & 31.86 & 35.51 & 33.15 & 19.68 & 43.78 \\
            \cmidrule{2-16}
            & LaCo  & 34.43 & 55.69 & 69.80 & 36.14 & \textbf{40.38} & 25.00 & 45.70 & 64.07 & 26.45 & 25.24 & 22.61 & 23.61 & \textbf{15.64} & \textbf{39.67} \\
            \cmidrule{2-16}
            & \makecell[c]{LaCo \\ +post train} & \textbf{34.92} & \textbf{61.88} & \textbf{73.18} & \textbf{38.12} & 36.54 & \textbf{34.62} & \textbf{57.49} & \textbf{66.21} & \textbf{29.47} & \textbf{25.33} & \textbf{28.33} & \textbf{29.87} & 10.02 & 38.58  \\
            \cmidrule{2-16}
             & \makecell[c]{ LaCo \\ +post train \\ +re prune}  & 33.80 & 45.35 & 65.07 & 23.27 & 36.54 & 0.96 & 38.49 & 60.43 & 26.07 & 25.37 & 23.07 & 22.98 & 15.48 & 36.71 \\
            
    \midrule
    \multirow{4}{*}{\makecell[c]{\textbf{Baichuan2} \\ \textbf{-7B}}} & Dense$^{*}$ & 33.37 & 67.56 & 76.17 & 82.67 & 41.35 & 63.46 & 63.14 & 63.30 & 54.25 & 56.95 & 52.63 & 51.04 & 20.84 & 64.55 \\
            \cmidrule{2-16}
            & LaCo & \textbf{33.00} & 52.28 & 68.50 & 76.24 & \textbf{42.31} & \textbf{26.92} & 47.26 & 56.15 & \textbf{31.53} & \textbf{31.24} & \textbf{28.99} & \textbf{27.72} & 12.03 & 50.85 \\
            \cmidrule{2-16}
            & \makecell[c]{LaCo \\ +post train} & 32.92 & \textbf{52.67} & \textbf{69.42} & \textbf{78.22} & 40.38 & 3.85 & \textbf{52.01} & \textbf{55.93} & 28.72 & 27.25 & 25.01 & 26.25 & \textbf{15.82} & \textbf{58.03}  \\
    \midrule
    \bottomrule
    \end{tabular}}
    \caption{The detailed scores across all benchmarks of pruned models, post-trained models, as well as post-trained models followed by re-pruning.}
    \label{tab:post_train_res}
\end{table*}

\begin{table*}[!t]
\setlength\tabcolsep{0.8pt} 
    \centering
    \resizebox{0.98\linewidth}{!}{
    \small
    \begin{tabular}{c|c|ccc|ccc|cc|cc|cccc}
    \toprule[0.5pt]
    \midrule
    \multirow{2}{*}{LLM} & \multirow{2}{*}{Ratio/Lay.} & \multicolumn{3}{c|}{Reasoning} & \multicolumn{3}{c|}{Language} & \multicolumn{2}{c|}{Knowledge} & \multicolumn{2}{c|}{Examination} & \multicolumn{4}{c}{Understanding} \\

    & & CMNLI & HeSw & PIQA & CHID & WSC$_{\text{P}}$ & WSC$_{\text{G}}$ & CSQA & BoolQ & MMLU & CMMLU &  Race$_{\text{H}}$ &  Race$_{\text{M}}$ & XSum & C3 \\
    \midrule
    \multirow{4}{*}{\makecell[c]{\textbf{Llama2} \\ \textbf{-7B}}} & 0\%/32 & 32.98 & 71.35 &  78.18 & 46.04 & 37.50 & 38.46 & 66.67 & 70.67 & 45.92 & 31.86 & 35.51 & 33.15 & 19.68 & 43.78 \\
            \cmidrule{2-16}
            & 12.0\%/28 & 32.99 & \textbf{55.91} & \textbf{74.48} & \textbf{42.57} & 36.54 & \textbf{29.81} & \textbf{52.58} & 60.12 & 25.59 & \textbf{27.10} & 22.01 & 21.73 & \textbf{17.97} & 36.44 \\
            \cmidrule{2-16}
            & 27.1\%/23 & \textbf{34.43} & 55.69 & 69.80 & 36.14 & \textbf{40.38} & 25.00 & 45.70 & \textbf{64.07} & \textbf{26.45} & 25.24 & \textbf{22.61} & \textbf{23.61} & 15.64 & \textbf{39.67} \\
            \cmidrule{2-16}
            & 45.0\%/17 & 32.58 & 38.33 & 60.07 & 20.30 & 36.54 & 0.96 & 34.73 & 61.38 & 23.98 & 25.59 & 22.38 & 23.26 & 1.28 & 38.85  \\
    \midrule
    \multirow{4}{*}{\makecell[c]{\textbf{Llama2} \\ \textbf{-13B}}} & 0\%/40 & 32.99 & 74.83 & 79.71 & 52.97 & 50.96 & 63.46 & 66.91 & 71.50 & 55.63 & 38.74 & 58.03 & 60.24 & 23.56 & 47.51 \\
            \cmidrule{2-16}
            & 14.6\%/34 & 32.99 & \textbf{71.88} & \textbf{76.82} & \textbf{51.98} & \textbf{63.46} & \textbf{48.08} & \textbf{63.72} & 63.43 & \textbf{53.97} & \textbf{38.23} & \textbf{59.35} & \textbf{61.49} & \textbf{21.32} & \textbf{47.67} \\
            \cmidrule{2-16}
             & 24.7\%/30 & 32.86 & 64.39 & 74.27 & 40.10 &  52.88 & 35.58 & 52.66 & \textbf{63.98} & 45.93 & 32.62 & 54.49 & 56.55 & 14.45 & 44.93 \\
            \cmidrule{2-16}
            & 49.7\%/20 & \textbf{34.22} & 46.55 & 63.82 & 13.37 & 56.73 & 10.58 & 36.28 & 62.23 & 38.41 & 27.24 & 51.97 & 56.41 & 1.56 & 36.38 \\
    \midrule
    \bottomrule
    \end{tabular}}
    \caption{The detailed results of models pruned at different pruning ratios using LaCo across all benchmarks. As the pruning ratio increases, overall model performance decreases. However, performance remains stable from 10-25\% pruning. Even at 50\% pruning, the model can maintain about 70\% performance.}
    \label{tab:varing_res_detailed}
\end{table*}

\end{document}